\relax
\documentclass[letterpaper]{article}
\usepackage{arxiv}
\usepackage{times}
\usepackage{helvet}
\usepackage{courier}
\usepackage[hyphens]{url}
\usepackage{graphicx}
\usepackage{graphicx}
\usepackage{booktabs}       
\usepackage{amsfonts}       
\usepackage{nicefrac}       
\usepackage{amsmath}
\usepackage{enumitem}

\begin{document}
\title{DNNs as Layers of Cooperating Classifiers}

\author{
  Marelie H. Davel, Marthinus W. Theunissen, Arnold M. Pretorius, Etienne Barnard\\
    Multilingual Speech Technologies, North-West University, South Africa; 
    and CAIR, South Africa. \\
  \{marelie.davel, tiantheunissen, arnold.m.pretorius, etienne.barnard\}@gmail.com \\
}

\maketitle

\begin{abstract}
A robust theoretical framework that can describe and predict the generalization ability of deep neural networks (DNNs) in general circumstances remains elusive. Classical attempts have produced complexity metrics that rely heavily on global measures of compactness and capacity with little investigation into the effects of sub-component collaboration.
We demonstrate intriguing regularities in the activation patterns of the hidden nodes within fully-connected feedforward networks. By tracing the origin of these patterns, we show how such networks can be viewed as the combination of two information processing systems: one continuous and one discrete. We describe how these two systems arise naturally from the gradient-based optimization process, 
and demonstrate the classification ability of the two systems, individually and in collaboration. This perspective on DNN classification offers a novel way to think about generalization, in which different subsets of the training data are used to train distinct classifiers; those classifiers are then combined to perform the classification task, and their consistency is crucial for accurate classification.
\end{abstract}


\section{Introduction}
\label{intro}

One of the central tenets of computational learning theory (CLT) is that the ability of a machine-learning system to generalize to unseen data results from its compactness. That is, if the system employs a number of parameters that is small relative to the number of training samples that it processes appropriately, we can be confident that the system will generalize well to unseen samples drawn from the same distribution as the training data. 

Several observations in recent years have raised questions about the applicability of this explanation in systems such as deep neural networks (DNNs). Most strikingly, Zhang et al.~\cite{zhang2016understanding} showed a number of cases where networks with very large capacity achieve excellent generalization performance. Although this work lead to a flurry of activity~\cite{shwartz2017opening,bartlett2017spectrally,neyshabur2017exploring,dinh2017sharpminima} and some controversy, it actually confirms long-observed weaknesses in the classical CLT bounds: going back to at least 1992~\cite{cohn1992tight}, it has been noted that those bounds are often so conservative as to not be useful in practice. It should also be noted that while parametric compactness is a sufficient condition for generalization, it has never been shown to be a {\it necessary} condition~\cite{kenji2019generalization}. Hence, the widespread search for a definition of model complexity that renders CLT applicable to DNN-like classifiers may in the long run prove fruitless.

In the current work, we investigate the capabilities of DNNs by studying the behavior of hidden nodes in some detail, limiting our attention to the conceptually simplest case of fully-connected feedforward classification networks with ReLU activation functions. We show that intriguing regularities in the activation patterns of nodes within such networks exist; and can be understood by analyzing the DNN training process as an interaction between two processes: one discrete and descriptive of the input patterns that a node is responsive to, and the other continuous and concerned with the magnitude of activation.
We verify that either of these processes can be used as basis for deriving node-based classifiers from a trained network. These observations suggest a novel way of viewing the behavior of DNNs as layers of cooperating classifiers. Although we do not directly relate this point of view to their generalization capabilities, our work suggests some novel perspectives that may contribute to such an understanding.

\section{An unexpected observation on node behavior}
\label{observation}

As motivated in Section \ref{intro}, we wish to understand the role of the hidden nodes within a trained DNN. By design, each of the output nodes corresponds to class membership, whereas each of the input nodes responds to a particular feature (and is therefore quite agnostic about class membership). Also, in a feedforward network without skipped connections, each layer of node activations is a comprehensive summary or ``state''~\cite{jiang2018margindistributions}: taken together, a layer of activations fully determines the activations in each of the subsequent layers.

\begin{figure}[t]
  \center{
  \includegraphics[width=0.40\textwidth]{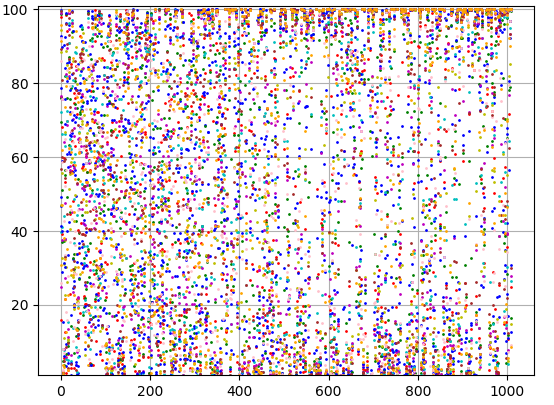}
  \includegraphics[width=0.40\textwidth]{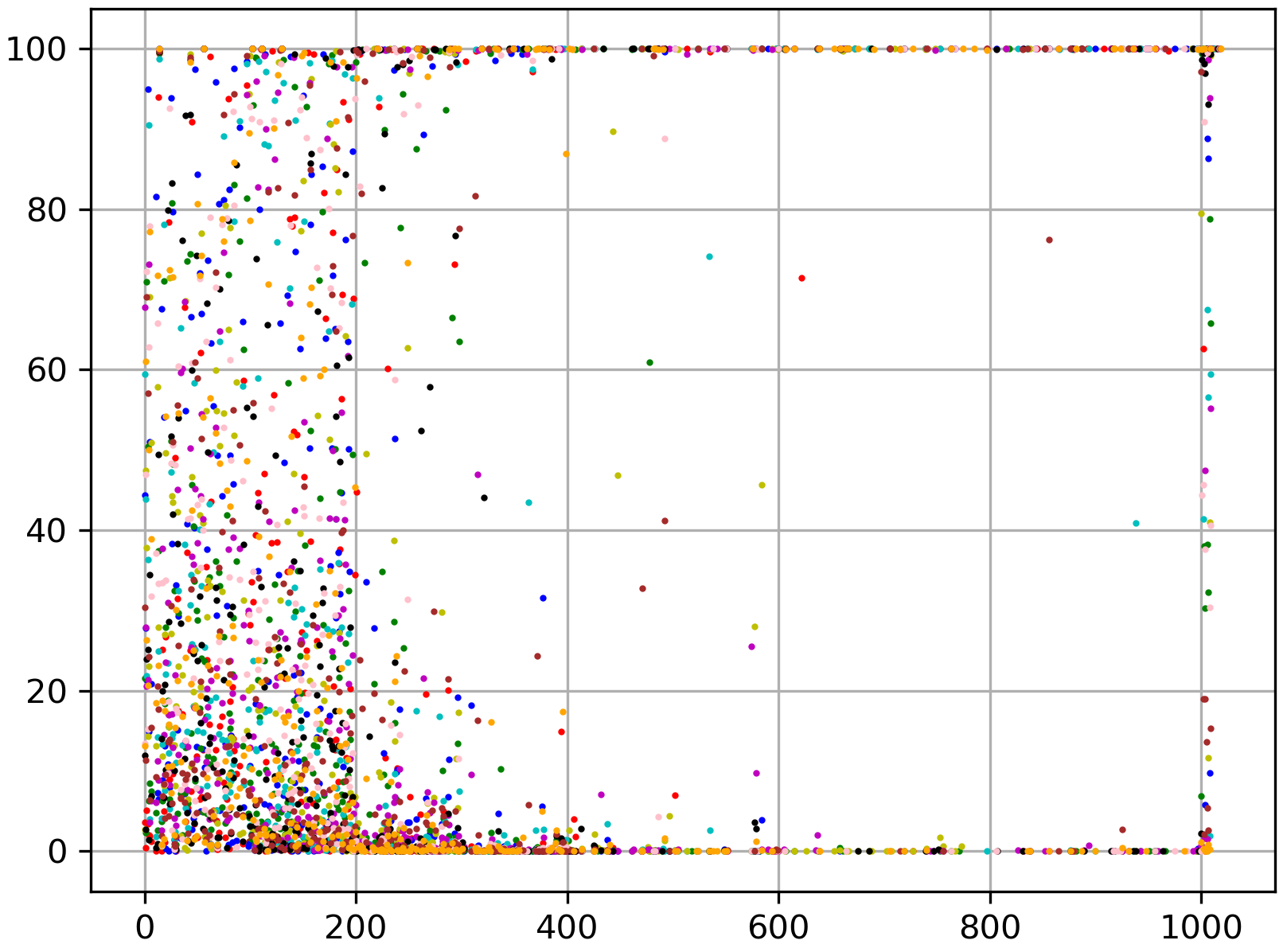}}
  \caption{Percentage of class samples that activate each hidden and output node of a network with 10 hidden layers and 100 nodes per layer, after initialization (left) and once trained (right) for MNIST digit recognition. Each class is indicated with a different color, and the nodes are ordered from input to output on the horizontal axis -- that is, nodes 0 - 99 correspond to the first hidden layer, 100 - 199 form the next hidden layer, etc. The final 10 nodes (after index 1000) are in the output layer.}
  \label{fig:node_class_activities}
\end{figure}

In a ReLU network, where a node is either activated or not, one can approach this question by asking how responsive each node is to inputs belonging to the different classes. Figure~\ref{fig:node_class_activities} shows an example of the activation patterns that we have observed in numerous ReLU-activated networks of various architectures, trained with different algorithms on different classification tasks. 

We observe that nodes in the first few layers are neither highly specific nor sensitive to any particular class: most nodes in the first two hidden layers are activated by {\emph some} samples from several classes. Deeper in the network, however, the nodes become highly selective: each node is activated by either none of the samples in a class or virtually all the samples in the class. This regular pattern occurs over a wide range of conditions, as long as the network has sufficiently many layers and nodes, and arises despite the random initialization of weights. It therefore seems to indicate a fundamental aspect of the way a DNN arranges itself to perform classification, and calls for an explanation in terms of the DNN training process. 

Earlier work on the complexity analysis of DNNs~\cite{Montfar2014OnTN,pmlr-v70-raghu17a,eldan2016power} has observed that hidden units in deeper layers produce many additional distinct linear regions in feature space; and that with depth, layer behavior becomes more abstract and class-specific.
However, the observed transition with depth is strikingly sharp, and not spread out over available depth as one would expect.
Below, we first introduce a measure that makes it easier to quantify the transition from class-agnostic to class-selective nodes, and then proceed with an analysis that investigates its genesis during gradient-based training.

\section{Layer Perplexity}
\label{perplexity}

Insight about the discrete dynamics of DNN training can be gained by investigating the number of different binary activation patterns (from here referred to as {\it patterns}) that occurs at each hidden layer. Each pattern consists of a vector of binary values indicating whether each node in the layer is active for a given input sample, or not. If the total number of occurrences of each pattern for a layer $l$ as a response to all samples from a class $c$ is given by the set $K(c, l)$, then the entropy of the patterns for class $c$ at layer $l$ can be defined as
\begin{equation}
H(c, l) = -\sum\limits_{n \in K(c, l)} \frac{n}{N_{c}} ln \biggl( \frac{n}{N_{c}} \biggr),
\label{eq:layer_entropy}
\end{equation}
where $N_{c}$ is the total number of samples belonging to class $c$; and the perplexity of the class $c$ at layer $l$ is defined as
\begin{equation}
P(c,l) = e^{H(c, l)}
\label{eq:layer_perplexity}
\end{equation}
In this context, entropy defines the average information content in the set of possible patterns and their frequencies, and  perplexity
provides an estimate of the total amount of information related to the patterns used by layer $l$ to represent all the samples in class $c$. Minimal information is indicated by a perplexity of $1$, which implies that the layer represents every sample of the class as an identical pattern. Maximal information is indicated by a perplexity value equal to the total number of samples in the class: this happens when every sample is represented by a unique pattern at the current layer.

\subsection{Trained models} 
\label{trained_models}

We conduct our experiments in a relatively simple setup. Our aim is to understand trends, 
while retaining the key elements that are likely to be common to high-performance DNNs. Thus, we use only fully-connected feedforward networks with highly regular topologies, and investigate their behavior on two widely-used image-recognition tasks, namely MNIST~\cite{Lecun98gradient-basedlearning} and FMNIST~\cite{xiao2017fashion}.
No data augmentation is employed.
Refinements such as drop-out and batch normalization are also avoided in order to focus on the essential mechanisms of DNN learning. (Such refinements do not contribute much to test set accuracy in this setting, in contrast to data augmentation, which does~\cite{ciresan2010deepbig,simard03bestpractices}.)

\begin{figure}[bp]
     \centering
     \includegraphics[width=0.43\linewidth]{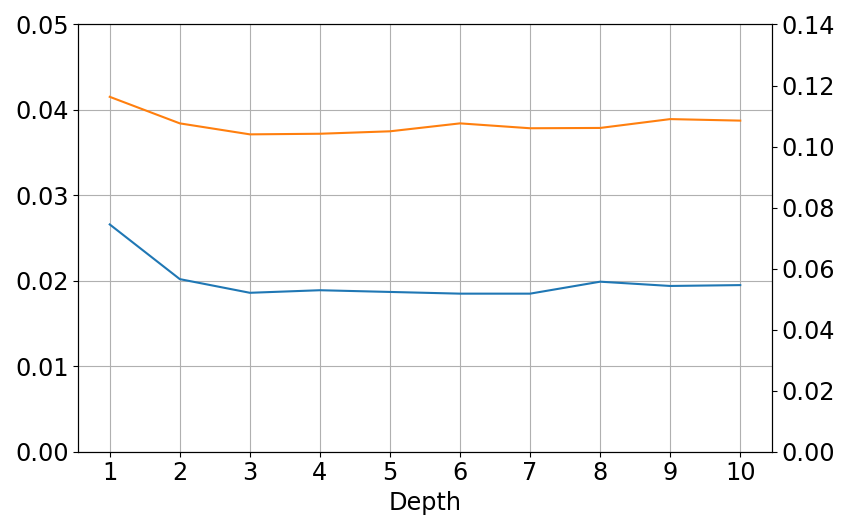}
     \includegraphics[width=0.43\linewidth]{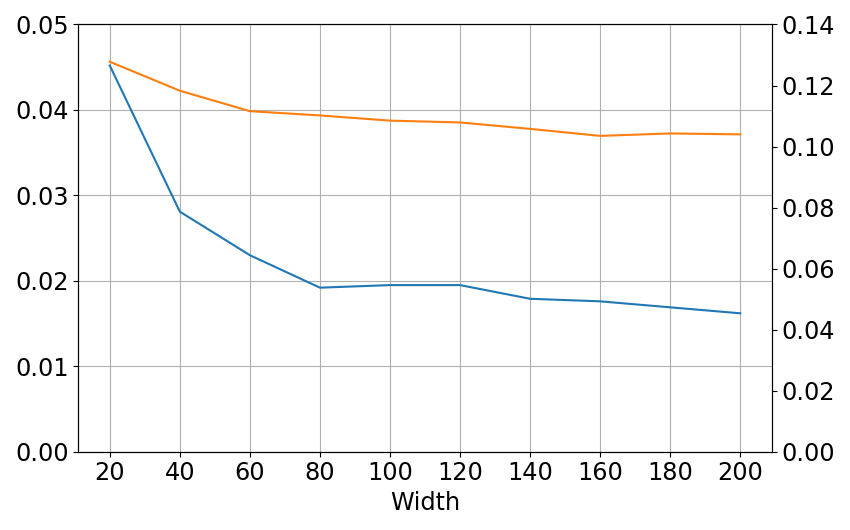}
     \caption{Test error for networks with varying depth and a width of 100 nodes (left) and varying width and a depth of 10 layers (right) trained on MNIST (blue curve, left vertical axis) and FMNIST (orange curve, right vertical axis).}
     \label{fig:network_performance}
\end{figure}

\begin{figure*}[tp]
	\center
	\includegraphics[width=0.9\linewidth]{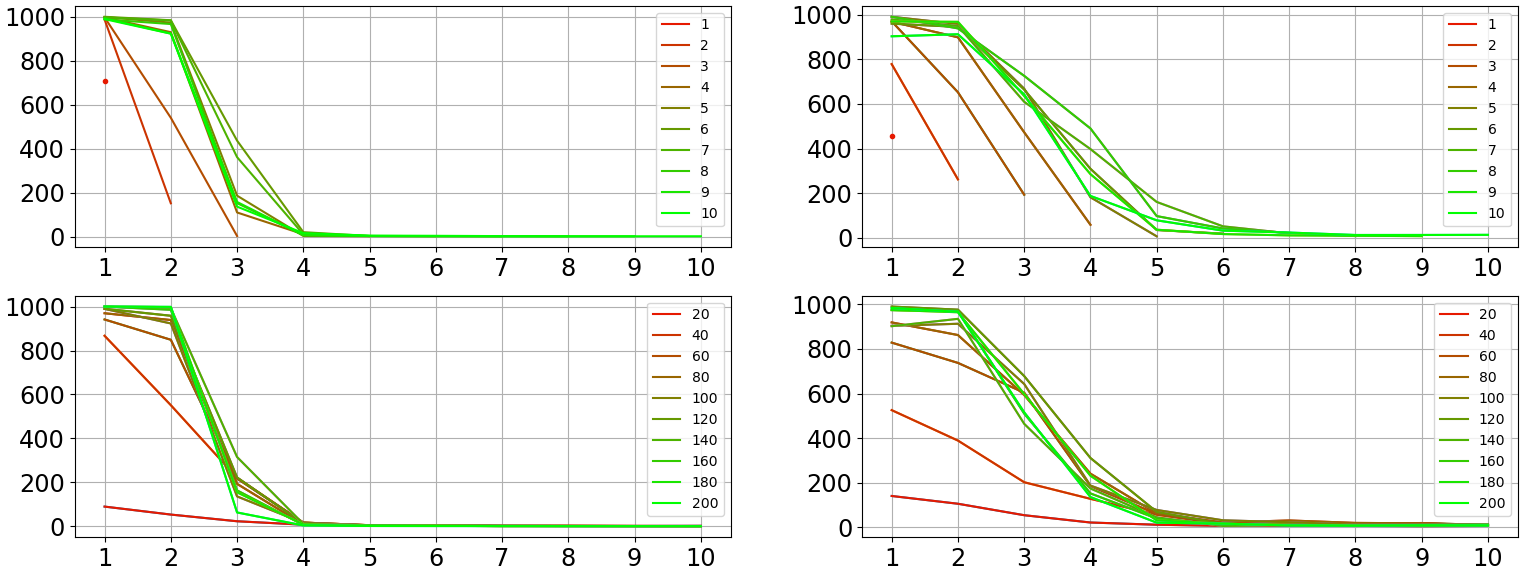}
  	\caption{Per-layer mean perplexity values with changing depth (top) and changing width (bottom) for MNIST (left) and {FMNIST} (right).}
  	\label{fig:perplexity}
\end{figure*}

All hidden nodes have Rectified Linear Unit (ReLU) activation functions, and a standard mean squared error (MSE) loss function is employed, unless stated otherwise. The popular Adam~\cite{kingma2014adam} optimizer is used to train the networks after normalized uniform initialization with three different training seeds~\cite{lecun2012efficient}, and the global learning rates are manually adjusted to ensure training set convergence. This is verified by ensuring that the performance obtained is comparable with prior results reported on both MNIST~\cite{Lecun98gradient-basedlearning,simard03bestpractices} and FMNIST~\cite{novak2018sensitivity,agarap2018deep}, where similar topologies were employed. We implement early stopping by choosing networks with the smallest validation error.

The performance of the trained models are shown in Figure~\ref{fig:network_performance}.
Our first analysis investigates several networks of fixed width and increasing depth. ``Depth'' here refers to the number of hidden layers, without counting the input or output layers. For a width of 100 nodes per layer, both the MNIST and FMNIST systems initially achieve decreasing error rates as the number of hidden layers grows, but the performance quickly saturates. 
However, increasing the number of layers (and thus parameters) beyond this level does not {\it degrade} performance, even when as many as 20 hidden layers are employed. (Results here shown up to a a depth of 10.)
In the second analysis, network depth is kept constant at 10 layers, and the width (number of nodes per layer) is adjusted. 
As with increased depth, increased width leads to a similar saturation in performance.

\subsection{Perplexity results} 
\label{perplexity_results}

Using the trained networks from Figure~\ref{fig:network_performance}, we now analyse the distinctiveness of their activation patterns. The per-layer mean perplexity of each network is shown in Figure~\ref{fig:perplexity}, with mean values obtained by averaging over all classes.  The perplexities are measured with regard to the {\it test set} samples: with the focus of our analyses being on generalization, we are interested in encodings that are applicable during categorization of unseen samples, and not only those created for optimization purposes. 

There are several interesting observations that can be made from these graphs. Notice the relatively sharp drop in perplexity values for all networks, and take note that the drop is more gradual for the FMNIST models and sharper for wider networks. Additionally, for the networks with sufficient depth and width: 
\begin{itemize}
    \item The perplexity in the later layers is very near 1. That is, all activation patterns have become fully class-specific.
    \item The perplexity values in the first 2 layers are almost equal to the total number of samples for each class (approximately 1,000 for the test sets for both MNIST and FMNIST), which means that an individual encoding is created per sample.
    \item The transition from high perplexity to low perplexity is very similar across networks.
\end{itemize}

\noindent
Lastly, notice that if the network width is below some threshold, the perplexity values in the earlier layers reduce accordingly. This cannot be due to a lack of representational power seeing as even the smallest layer (20 nodes) can represent more patterns than required by the number of training samples. Taking into account their lower test error, this suggests that wider networks represent sample information in a way that is more conducive to good generalization. This phenomenon was recently explored by \cite{pmlr-v97-brutzkus19b} where it is attributed to better weight exploration and a small number of observed prototype weight vectors.

\subsection{Discussion}
Provided the network is large enough, there seems to be a range of earlier layers within which the nodes have high (virtually maximal) perplexity, and a corresponding range of later layers where the nodes have relatively low (virtually minimal) perplexity. Furthermore, the transition from the former behavior to the latter is consistent across all networks, irrespective of size, as long as they are deeper and wider than a task-specific threshold. After this transition, the class-specific discrete behavior in the excess layers is relatively trivial. (Perplexity is already at a minimum.)

The nodes in the earlier layers appear to perform most of the information processing required to produce a feature space that supports the ability to differentiate among samples relating to different classes. In this setup, the deeper layers effectively produce no new benefits and merely propagate the information forward through the network. The forward propagation of information, at this point, takes the form of a class-specific encoding, which is unique to each layer.
By varying either width or depth, the same message emerges: a task-specific threshold exists with regard to both width and depth, beyond which network behavior is strikingly regular and similar, irrespective of network size.

\section{Theoretical perspective} 
\label{theory}

In Section \ref{perplexity} it was shown that, once trained, a ReLU-activated multilayer perceptron (MLP) exhibits behavior that is clearly discrete: the activation patterns of each layer display distinct encodings, closely related to sample encodings in the first layer, and class encodings in the later layers.
In this section, we analyze the training process in order to determine how the stochastic gradient descent (SGD) equations give rise to this discrete behavior.
The MLP we study is allowed an arbitrary number of layers and nodes per layer, with each layer fully specified by its weight matrix. Initially we consider an arbitrary loss function but then restrict the analysis to mean squared error (MSE) and cross-entropy (CE) loss, using matching activations in the output layer (linear or softmax, respectively).  
We use $w_{i,j,k}$ to denote the individual weight from node $k$ in layer $i-1$ to node $j$ in layer $i$. Bias is dealt with as an extra weight in the first layer only, associated with an extra feature of value 1. (Given sufficient width, a bias node is not necessary beyond the first layer of an MLP.)

\subsection{Gradient-based optimization}

Gradient-based optimization has many variations but is essentially a straightforward process. In its basic form, each weight update is accumulated over a batch of random samples, each sample contributing a $\Delta w_{i,j,k}$. 
Each sample-specific update is proportional to the derivative of the error function $E$ with regard to this weight, and the learning rate $\eta$ (which could potentially be adaptive, as with Adam). In practice, the derivative of the error function with regard to each parameter is calculated using backpropagation
\begin{eqnarray}
\Delta w_{i,j,k} = - \eta\frac{\partial{E}}{\partial{w_{i,j,k}}} = - \eta\beta_{i,j}a_{i-1,k}
\label{eq:backprop}
\end{eqnarray}
with $a_{i-1,k}$ the activation result at layer $i-1$ for node $k$ and $\beta_{i,j}$ as defined below. Using $z_{i,j}$ to describe the sum of the input to node $j$ in layer $i$, and defining the symbols
\begin{eqnarray}
\alpha_{i,j}  =  \frac{\partial a_{i,j}}{\partial z_{i,j}} \label{eq:alpha_ij} \hspace{5mm}
\lambda_{j}  =  \frac{\partial{E}}{\partial{a_{N,j}}} \label{eq:lambda_j}
\end{eqnarray}
$\beta_{i,j}$ is calculated by counting through all $n$ forward connections from node $j$ to the next layer, working backwards from the last layer (also counting the output layer) $N$:
\begin{eqnarray}
\beta_{i,j} = \begin{cases}
	\alpha_{i,j} \sum\limits_{n} w_{i+1,n,j} \beta_{i+1,n}
	& \text{if } i \neq N \\
    \alpha_{i,j} \lambda_j
    & \text{if } i = N
\end{cases}
\label{eq:beta_ij}
\end{eqnarray}
This recursive update rule is important for computational efficiency but, while not commonly done, the derivative can also be written as an iterative expression\footnote{See Appendix \ref{app:derivations}.}:
\begin{eqnarray}
\beta_{i,j} 
& = & \sum\limits_{b=0}^{B_i-1} \lambda_{I_{i,j}(N,b)}\prod\limits_{g=i}^N \alpha_{g, I_{i,j}(g,b)} 
\prod\limits_{r=i+1}^{N} w_{r, I_{i.j}(r,b), I_{i,j}(r-1,b)} \label{eq:beta_iter}  \\
\textrm{where } B_L & = & 
\begin{cases}
\prod\limits_{m=L+1}^{N} s_m & if L \neq N \notag \\
1 & if L = N 
\end{cases} \notag \\
I_{i,j}(r,b) & = & 
\begin{cases}
(b \div B_r) \mod s_r & \textrm{ if } r \neq i \notag \\
j & \textrm{ if } r = i 
\end{cases}
\end{eqnarray}
with $s_i$ the number of nodes in layer $i$, and each $I_{i,j}(r,b)$ indexing function specific to the layer and node position of the $\beta_{i,j}$ required.
When inner node activations are ReLUs, this equation simplifies further. Noting that 
\begin{eqnarray}
Relu(x) & = & x T(x) \\
\textrm{where }T(x) & = & \begin{cases}
    1		\text{ if } x > 0 \\
    0       \text{ if } x \leq 0 \\ 
\end{cases}
\end{eqnarray}
the weight update becomes
\begin{eqnarray}
\Delta w_{i,j,k}
= -\eta \hspace{1mm} a_{i-1,k} \sum\limits_{b=0}^{B_i-1}& \lambda_{I_{i,j}(N,b)}
\prod\limits_{g=i}^{N-1} T(z_{g, I_{i,j}(g,b)}) \notag \\
&\prod\limits_{r=i+1}^N w_{r, I_{i,j}(r,b), I_{i,j}(r-1,b)} \label{eq:relu_delta_w_ijk}
\end{eqnarray}
In effect, the $b$ index runs through all possible paths from node $j$ in layer $i$ to each of the nodes in layer $N$, the $g$ index runs through all the activation values of a single path, and the $r$ index multiplies the weights along the same path. 
Using MSE as loss function and linear activation functions in the outer layer results in $\lambda_{i,j} = z_{N, I_{i,j}(N,b)} - y_{I_{i,j}(N,b)}$ where $y_j$ the true target value at the outer node $j$, that is, the classification gap. Note that  $\lambda_{i,j}$ has the same form when using a cross entropy loss function with softmax activation functions in the outer layer, as long as one-hot encodings are used for classification targets.

Per sample, each weight update then only takes into account the activation strength at node $k$ feeding into the weight, and all the active paths -- where all the $T(.)$ values are 1 -- supported from node $j$ onward. Each path contributes a single product of all the weights along the active path, multiplied by the classification gap at the path end point. The $T(.)$ values can therefore by viewed as switches, selecting which samples contribute to a weight update at each point in the network, and the weight update rewritten as:
\begin{equation}
\Delta w_{i,j,k}
 = \eta \hspace{1mm} 
\sum\limits_{s \in S } 
\sum\limits_{p \in P_s}
(a^s_{i-1,k} )( \prod\limits_{g=1}^{N-i} w'_{p_g} ) (y^s-z^s_{N,p_{N-i}})
\label{eq:simplified_sum}
\end{equation}
where $S$ consists of all the samples active at both nodes $j$ and $k$, $P_s$ is the set of active paths that starts at node $j$ (generated specifically by $s$) and $w'_{p_g}$ runs through the $g$ weights along the active path $p = p_1, p_2, \ldots p_{N-i}$. The $s$ superscript emphasizes that these are sample-specific values.

\subsection{Two collaborative systems}
\label{collaborative_systems}

The update process of Equation~\ref{eq:simplified_sum} can be viewed as two interacting systems: one continuous and one discrete, both utilizing the same underlying network architecture and parameters. Each node plays a role in both systems: 
\begin{enumerate}
    \item The discrete system associates an ``on/off'' value with every single sample-node pair, depending on whether the node is active or not for that sample. This system is fully specified by the $T(.)$ values of Equation~\ref{eq:relu_delta_w_ijk}. Nodes can therefore be considered as switched either on or off, giving rise to a discrete information processing system that creates a discrete set of samples at each node.
    \item The continuous system associates a continuous value with each sample-node pair (the pre-activation value of the sample at the given node) and updates the continuous values of the weight vector feeding into this node during gradient descent.
\end{enumerate} 
The training process utilizes both systems to optimize the network, but the relative importance of the two systems with regard to eventual classification ability changes, both during the training process and through the layers of the network. 
Each node in effect acts as a local feature transformation, combining multiple features from an earlier level to form a single new feature, made available to the next level. The node only optimizes its weights (weights feeding into the node) with regard to the set of samples it is sensitive to: with regard to these, it determines the relative importance of the features available at the previous layer in closing the classification gap it is aware of.
The training process uses the two systems interactively:
(1) During the forward pass, the discrete systems determines whether a sample should be included or excluded from the set of similar samples at that node. (2) During the backward pass, only the selected samples are used by the continuous system to update the relative weighting of the input features: creating a new feature more attuned {\it to these specific samples}, and these only. 

This also means that the optimization process is simultaneously taking into account both global and local information. Globally, the extent to which all the collaborating nodes have already ``solved'' the task posed by a specific sample determines the influence of that sample, while locally, each node that is active for an unsolved sample adjusts its parameters according to its own set of active samples only. Locally, nodes solve subsets of the class differentiation task; globally, nodes in a layer cooperate. 

\section{Empirical confirmation for two systems}
\label{empirical_confirmation}

One way in which to determine the extent to which the discrete and continuous systems each exists in own right, is to analyze the classification ability of each system individually. 
We ask how well each system would be able to classify unseen samples, given either the discrete information available per sample (which nodes are on or off) or the continuous information per sample (pre-activation values at each node). 

\subsection{Nodes as classifiers} 
\label{node_classifiers}

We now interpret each node as a classifier, implicitly estimating $P(z|y_n)$, where $z$ is the pre-activation value and $y_n$ a class. A discrete, continuous and combined estimate of this value is created at each node:
\begin{itemize}
    \item {\it discrete}: if $z > 0$, $P(z|y_n)$ is estimated as the ratio of class $n$ training samples with positive activation values with regard to all class $n$ training samples; 1 minus this value otherwise. 
    \item {\it continuous}: the estimate provided by a kernel density estimator trained using all class $n$ training data activation values observed at this node.
    \item {\it combined}: using the discrete estimate if $z \leq 0$, the continuous estimate otherwise.
\end{itemize}
This estimate is combined with the prior probability $P(y_n)$ of a class being observed to estimate the posterior $P(y_n|z)$:
\begin{eqnarray}
P(y_n|z) & = & \frac{P(z|y_n)P(y_n)}{\sum_m{P(z|y_m)P(y_m)}} \label{P(y_n|z)}
\end{eqnarray}

We view the nodes as independent classifiers (we ignore possible dependence) and multiply the probability estimates per class 
over all the nodes in a layer, to obtain a layer-specific probability estimate for each of the three systems. (In practice, the log probabilities are summed.) These probability estimates can then be used directly to classify samples based on maximum probability, creating three layer-specific classifiers for each layer in the network: a continuous, a discrete and a combined classifier. 
While neither the nodes nor the layers use these probabilities directly, they provide insight into the information available locally at each point in the network. By evaluating layer-specific classification ability at different layers and at different stages in the training process, we can better demonstrate the interaction between the discrete and continuous systems.

\begin{figure*}[b!]
  \center{
  \includegraphics[width=0.90\textwidth]{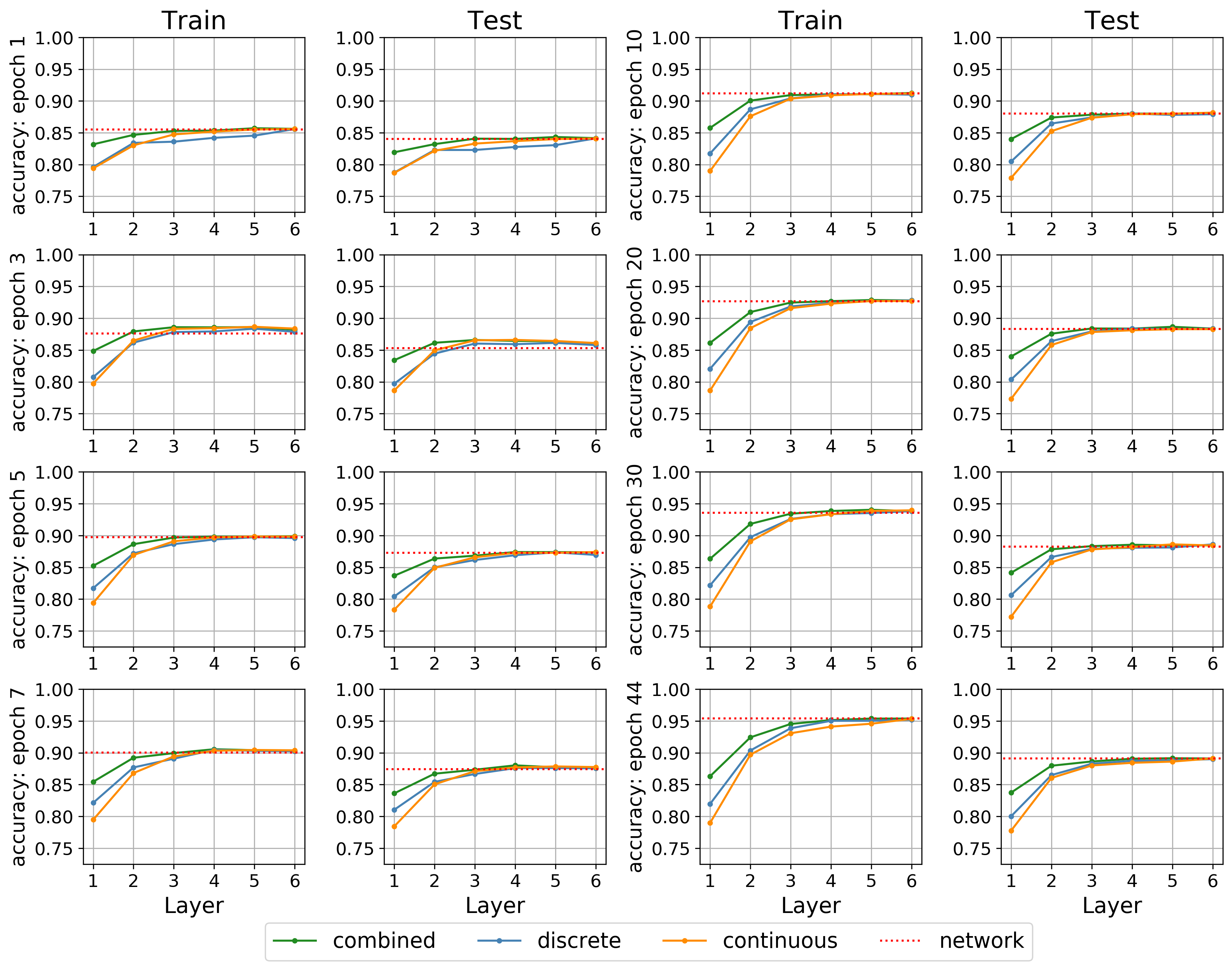}}
  \caption{Train and test accuracies of the discrete, continuous and combined systems as measured on an FMNIST 6x100 network. System performance is shown after specific epochs.
  The red dotted line (``network'') indicates the performance of the MLP itself when evaluated in the conventional manner. }
  \label{fig:6x100_epoch}
\end{figure*}

\subsection{Classification ability during training} \label{classification_ability}

\begin{figure}[tb]
  \center{
  \includegraphics[width=0.75\linewidth]{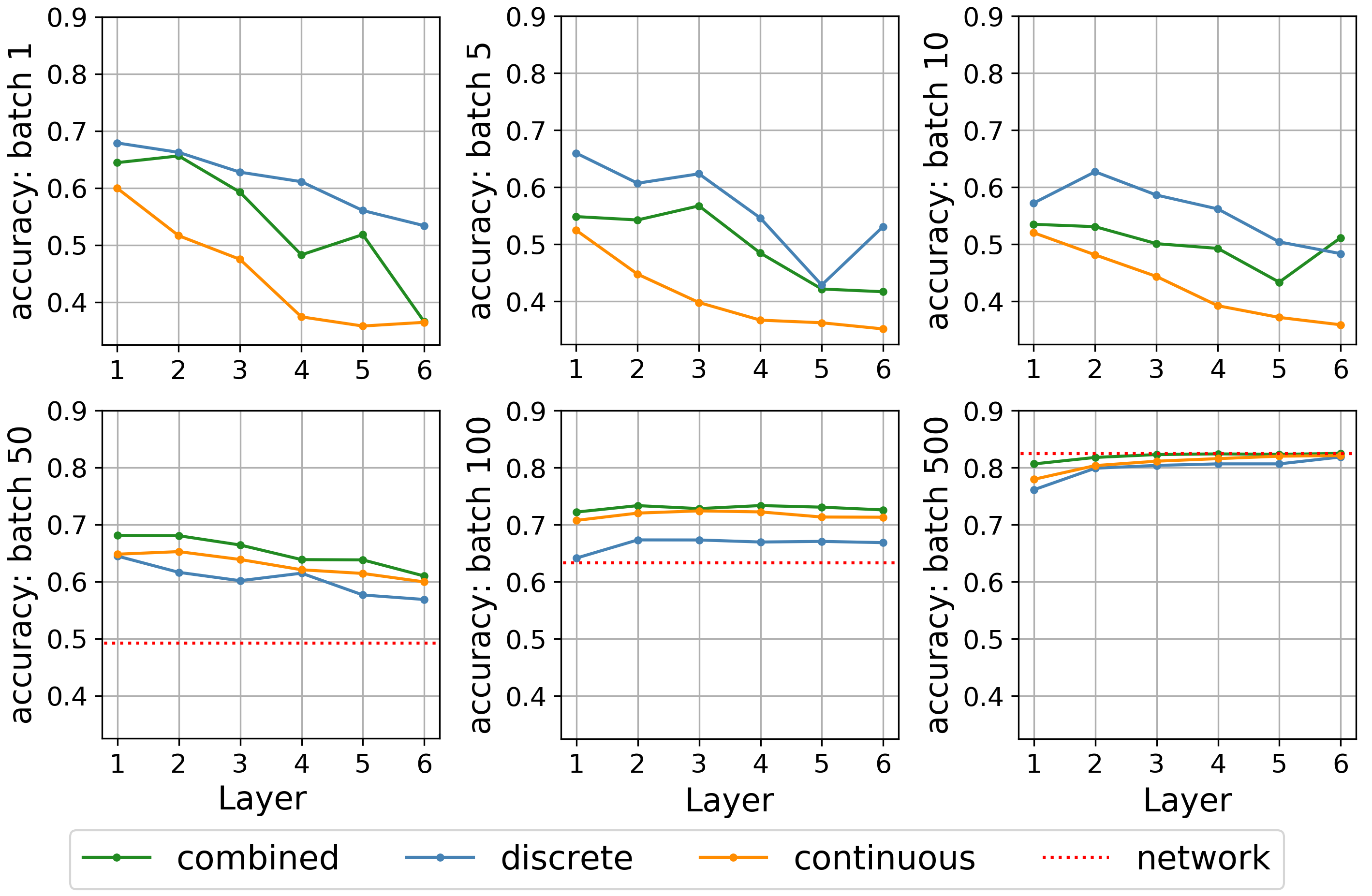}}
  \caption{The same analysis (for test data only) as in Figure \ref{fig:6x100_epoch}, except that results are not shown per epoch but after specific mini-batch 
  updates in the first epoch.}
  \label{fig:6x100_batch}
\end{figure}

Using the nodes as individual classifiers, we evaluate the performance of the discrete, continuous and combined systems generated from the trained models in Figure~\ref{fig:network_performance}, during the training process.
In Figure~\ref{fig:6x100_epoch}, we demonstrate the performance of an MLP
with $6$ hidden layers of $100$ nodes each, trained on the FMNIST classification dataset; the behavior of this model during training expresses the overall tendencies for all the analysed models very well. 

The most striking observation is that, at the later hidden layers, the accuracies of the three systems are virtually identical. In the first layer, the accuracy of the combined system is higher than both the discrete and continuous systems. This difference in classification accuracy among the three systems becomes smaller at later layers in the network, until it disappears. 
While it is to be expected that the combined system would outperform the other two (since its probability estimates have access to information pertaining to both the continuous and discrete subsystems) this is not what happens: at later layers, the other two systems are able to perform at levels comparable to the system subsuming them. 

Additionally, it can be seen that the accuracies in the later layers improve visibly over iterations of learning while the performance of the earlier layers improves less. This reinforces the idea that the function of earlier layers is not to classify samples into the classes involved in the global classification problem, but instead act as general sample differentiators (that is, earlier layers attempt to group and solve subsets of the main task, which may not necessarily be class-specific); later layers use these elements to more efficiently perform the classification task. 
During training, the overall accuracy of each system in later layers increases on the train and on the test set until it reaches the same, or slightly better accuracy as the network itself.
At the end of the first epoch, significant training has already occurred. We therefore also investigate how the performance of these systems changes during mini-batch updates in the first epoch, as shown in
Figure~\ref{fig:6x100_batch}. Note how poorly the continuous system performs initially (relative to the discrete system), until the training process stabilises and the previously discussed trends emerge.

\begin{figure}[t]
  \center{
  \includegraphics[width=0.8\linewidth]{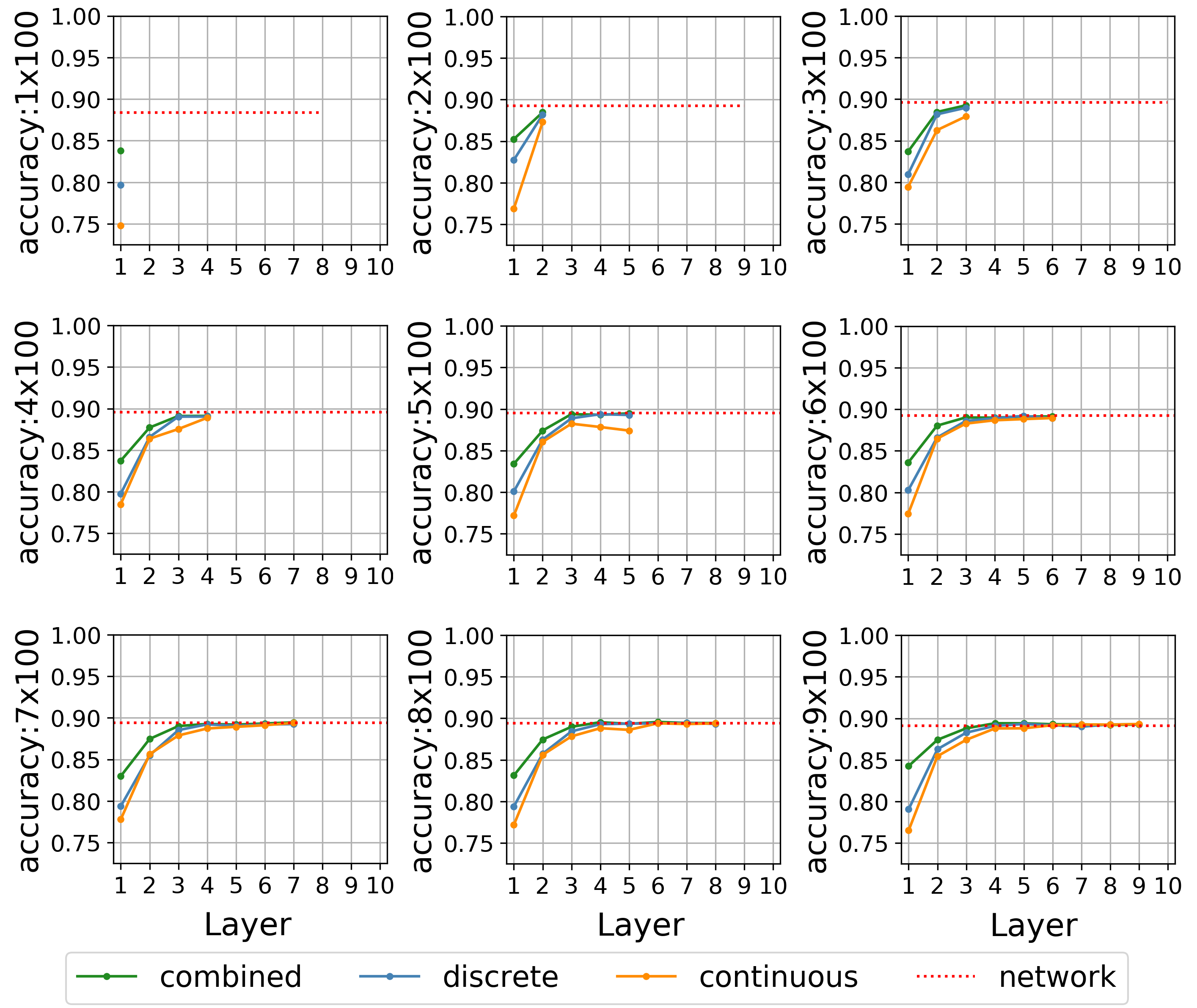}}
  \caption{Discrete, continuous and combined system test accuracies for networks with varied depth (1-9) (FMNIST).}
  \label{fig:var_x_100}
\end{figure}

Similar trends\footnote{Also see Appendix \ref{app:width}.} are observed when changing either network width or depth. 
Figure~\ref{fig:var_x_100} depicts the classification accuracy of the three systems for a set of FMNIST networks with fixed width (100 nodes) and increasing depth (1 to 9 layers). It is striking to note that the three systems start overlapping when sufficient depth becomes available, but struggle to beforehand. 
Similarly, when the network layers lack width, the  earlier layers underperform significantly. This is especially true for the discrete system. As expected, there is a clear increase in accuracy (across all systems) in the later layers with an increase in width. Curiously, the continuous performance appears to reduce with an increase in width in the first layers.

While not shown here, trends for FMNIST and MNIST 
are similar, except that for MNIST
(1) the depth at which the three systems converge is earlier; (2) higher accuracies are observed overall; 
and (3) there is an anomalously low performance measurement for the discrete system at one of the layers of the model with a width of 20. 
(We know that the discrete subsystem tends to underperform significantly at low widths.)
Finally, it is clear that the the nodes at each layer have the ability to solve the classification task when applied in collaboration. It is worth noting that, in the earlier layers, nodes are formed that range from very general (active for many samples) to very specific (active for only one or two samples).

\section{Alternative design choices}

The trends presented in this paper are based on the learning dynamics of an MLP using ReLU activation functions. This section briefly discusses to what extent the findings are applicable to deep learning models with alternative design choices, including activation functions that are not piece-wise linear. While we do not extend our analysis to more complex deep learning architectures, we do refer to related work where analogous observations were made with regards to other architectures. 

\begin{figure}[ht!]
  \center{
  \includegraphics[width=0.65\linewidth]{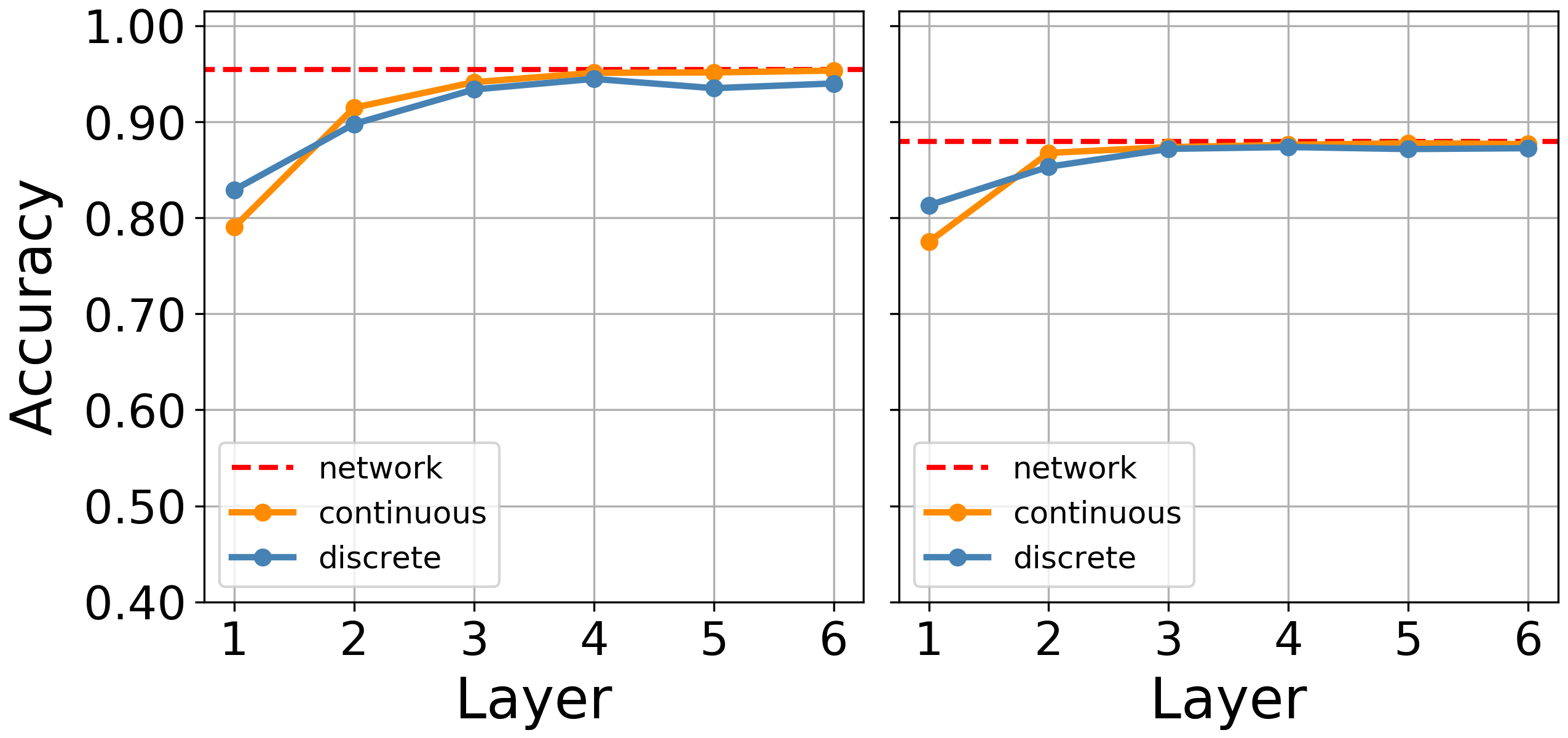}}
  \caption{Train and test accuracy of the discrete and continuous system in a 7x100 network using sigmoid activation functions (FMNIST).}
  \label{fig:sigmoid_depth}
\end{figure}

It is not too unexpected that ReLUs -- with their piece-wise linear characteristics -- would demonstrate discrete behavior, but what happens if the activation function has a continuous nature?
Specifically, we repeat the above two-system analysis using sigmoid activation functions instead of ReLUs.
This time we define the node as ``switched on'' for all activation values greater than 0.5 (and as `` switched off'' otherwise). 
Intuitively this choice makes sense, as this is the point at which the sigmoid function has maximal gradient and activation values are expected to diverge away from this value toward 0 or 1.
Somewhat surprisingly, the discrete system again emerges very clearly, as shown in Figure~\ref{fig:sigmoid_depth}, where classification performance is demonstrated for a 7x100 MLP that is similar to previous models, except that sigmoid activations and a CE loss function is used. 
We see that the two systems in the sigmoid-activated network behave similarly to those in the ReLU-activated networks, except that the continuous system outperforms the discrete system by a small margin in deeper layers. Other trends remain.

In addition, we empirically confirm that the trends discussed in Sections \ref{perplexity} and \ref{empirical_confirmation} are present in ReLU-activated MLPs with several alternative optimizers, loss functions, output functions, and classification data sets. We observed quantitative variations but no qualitative inconsistencies for the alternatives tested. We did find that choices that introduce a form of noise into the training process (such as batch normalization, explicit training data noise or non-adaptive optimizers) generally increase layer perplexities and reduce hidden unit saturation.

It has long been known that Convolutional Neural Network (CNN) layers
create feature spaces in a hierarchical structure, with earlier layers representing more general sample information and later layers becoming more specific, often thought of as a transition from local to global feature information~\cite{Zeiler2013VisualizingAU,Ma2015HierarchicalCF}.
In \cite{Alain2016UnderstandingIL} it was found that by training linear classifiers using the features produced by each layer in popular CNN models, such as Inception v3 and Resnet-50, one can estimate the utility (in terms of linear separability) of feature representations at each layer. Similarly, in \cite{Montavon2011KernelAO} kernel analysis was used to rate the representations produced by each layer in MLPs and CNNs according to their simplicity and power to predict classes accurately. While focused on layers as classifiers, rather than smaller elements (as we do), the results of both of the latter works are consistent with our own in that: (1) later feature spaces perform better than earlier ones, (2) the transition from general to class-related features is monotonic and surprisingly regular, and (3) the transition is more gradual for a task with more class variance and overlap. This suggests that some of our findings may be extendable to more complex, heavily engineered, deep learning architectures.

The heart of the results in this paper is based on the insight that weight vectors (fanning into a node) can be analyzed as isolated units,  each trained to reduce a portion of the global error in terms of a sub-population (within which the samples are inherently similar) of the training set, by utilizing either a hard (ReLU) or weighted membership rule. It is, therefore, very likely that such an analysis is applicable to other deep learning models built on the principle of updating weight vectors through gradient descent in conjunction with a non-linear activation function.

\section{Conclusion}

In this work we presented interesting regularities in the class-related activation patterns of nodes within a deep ReLU-activated network. We showed that fully-connected feedforward networks systematically ``compress'' their class discrimination into the early layers of a network, across a wide range of parameters and tasks. The origin of this behavior was studied through a theoretical investigation into the gradient-based optimization of such networks, highlighting the role of locally relevant nodes in solving the network-wide task. Specifically, nodes can be shown to create discrete clusters of samples that they are particularly attuned to. This phenomenon suggests that we investigate the discrete and continuous aspects of such networks separately, and we have shown that both discrete and continuous node-based probability estimators can be constructed to perform highly accurate layer-by-layer classification. 

Our analysis suggests that the generalization strength of DNNs arises from the collaborative contributions of the separate classifiers (some very general, some very specific) that are formed by individual nodes, and we are currently investigating how to quantify the properties of such distinct but collaborative units, which select variable sets of training samples to optimize their training set accuracy.

\bibliographystyle{unsrt}
\bibliography{references}

\appendix

\section{Additional derivations}
\label{app:derivations}

This section contains additional detail on the derivation of Equations \ref{eq:relu_delta_w_ijk} and \ref{eq:simplified_sum}, using the same notation as in Section 4.1.

\begin{figure}[hb]
  \center{
  \includegraphics[width=0.40\textwidth]{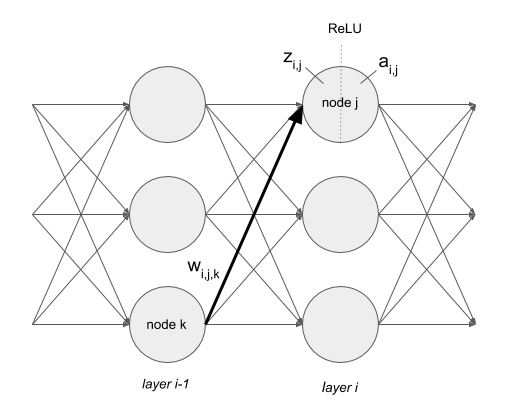}}
  \caption{Equations address the gradient descent update to the weight linking node $k$ to node $j$.}
  \label{fig:notation}
\end{figure}

\subsection{Derivation of equation \ref{eq:relu_delta_w_ijk}}
\label{app:relu_delta_w_ijk}

The $\beta_{i,j}$ from Equation \ref{eq:beta_ij}:
\begin{eqnarray}
\beta_{i,j} = \begin{cases}
	\alpha_{i,j} \sum\limits_{n} w_{i+1,n,j} \beta_{i+1,n}
	& \text{if } i \neq N \\
    \alpha_{i,j} \lambda_j
    & \text{if } i = N
\end{cases}
\end{eqnarray}
can be expressed as an iterative product of sums of products of sums, etc.
For $i \neq N$:
\begin{eqnarray}
\beta_{i,j} & = & 
\alpha_{i,j}\sum_{k_1=1}^{s_{i+1}} w_{i+1,k_1,j} 
\alpha_{i+1,k_1}\sum_{k_2=1}^{s_{i+2}} w_{i+2,k_2,k_1} 
\alpha_{i+2,k_2}\sum_{k_3=1}^{s_{i+3}} w_{i+3,k_3,k_2}  \notag \\ 
&& \ldots \alpha_{N-1,k_{N-i-1}}
\sum_{k_{N-i}=1}^{s_{N}} w_{N,k_{N-i},k_{N-i-1}} 
\alpha_{N,k_{N-i}}\lambda_{k_{N-i}}
\end{eqnarray}
where $n_l$ indicates the number of nodes in layer $l$.
Defining a helper indexing function that carefully enumerates the components:
\begin{eqnarray}
B_L & = & 
\begin{cases}
\prod\limits_{m=L+1}^{N} s_m & if L \neq N \notag \\
1 & if L = N 
\end{cases} \notag \\
I_{i,j}(r,b) & = & 
\begin{cases}
(b \div B_r) \mod s_r & \textrm{ if } r \neq i \notag \\
j & \textrm{ if } r = i 
\end{cases}
\end{eqnarray}
and using $I(r,b)$ as a shorthand for $I_{i,j}(r,b)$ when calculating a specific $\beta_{i,j}$, the expression can be rewritten as:
\begin{eqnarray}
\beta_{i,j} & = & \sum\limits_{b=0}^{B_i-1} 
\prod\limits_{g=i}^{N-1} \alpha_{g, I(g,b)} 
w_{g+1, I(g+1,b), I(g,b)} 
\alpha_{N, I(N,b)} \lambda_{I(N,b)} \notag \\
& = & \sum\limits_{b=0}^{B_i-1} \lambda_{I(N,b)}\prod\limits_{g=i}^N \alpha_{g, I(g,b)} 
\prod\limits_{r=i+1}^{N} w_{r, I(r,b), I(r-1,b)}
\label{eq:beta_ij_derivation}
\end{eqnarray}
When $i=N$, then $B_N = 1$, $\lambda_{I_{N,j}{N,0} = \lambda_j}$
and $\beta_{N,j} = \lambda_j\alpha_{N,j}$, which means that Equation \ref{eq:beta_ij_derivation} holds for all $i \leq N$. 

In the case of a ReLU-activated network, the derivative of the post-activation value with regard to the pre-activation value is either 1 or 0, depending on the pre-activation value, that is, $\alpha_{a, b} = T(z_{a,b})$, with $T$ as defined in Equation 8. 
This leads directly to Equation \ref{eq:relu_delta_w_ijk}:
\begin{eqnarray}
\Delta w_{i,j,k}
 = -\eta \hspace{1mm} a_{i-1,k} \beta_{i,j}\notag \\
= -\eta \hspace{1mm} a_{i-1,k} \sum\limits_{b=0}^{B_i-1}& \lambda_{I(N,b)}
\prod\limits_{g=i}^{N-1} T(z_{g, I(g,b)}) 
\prod\limits_{r=i+1}^N w_{r, I(r,b), I(r-1,b)} \notag
\end{eqnarray}

\subsection{Derivation of Equation \ref{eq:simplified_sum}}
\label{app:simplified_sum}

As the derivative $\lambda_{j}$ of an MSE loss function with linear output nodes is simply $z_{N,j}-y_j$ at each node $j$, the update rule becomes:
\begin{eqnarray}
\Delta w_{i,j,k} & = & \eta  \hspace{1mm} a_{i-1,k} \sum\limits_{b=0}^{B_i-1} 
\hspace{1mm}  (y_{I(N,b)}-z_{N,I(N,b)})
\hspace{1mm} \prod\limits_{g=i}^{N-1} T(z_{g, I(g,b)}) 
\prod\limits_{r=i+1}^N w_{r, I(r,b), I(r-1,b)} 
\label{eq:update_rule_app}
\end{eqnarray}

Note that a similar result can be derived if softmax output nodes are combined with a cross-entropy loss function.
The derivative of a softmax function $(p_k)$ at one node $k$ is influenced by the values at all other output nodes, that is: 
\begin{eqnarray}
\frac{\partial{p_k}}{\partial{p_j}} = \begin{cases}
p_k(1-p_k) & if j=k \\
- p_jp_k & if j \neq k
\end{cases}
\end{eqnarray}
Using softmax to produce the probability estimate $p_k$ at each node $k$ (of the node being active, given the training sample), and using the cross-entropy loss function ($L$) to estimate the error between target vector $\overline{y}$ and output vector $\overline{z_{N}}$ will then give:
\begin{eqnarray}
L & = & - \sum\limits_{k} y_k \log{p_k} \notag \\
& = & - \sum\limits_{k} y_k \frac{1}{p_k}\frac{\partial{p_k}}{\partial{p_j}} \notag \\
& = & - \sum\limits_{k\neq j} [y_k \frac{1}{p_k}(-1)(p_k)(p_j)]
- y_j \frac{1}{p_j}(p_j)(1-p_j) \notag \\
& = & p_j\sum\limits_k 
y_k - y_j \notag \\
& = & z_{N,j} - y_j \textrm{ if }\sum\limits_k y_k = 1
\end{eqnarray}
that is, if assuming one-hot encoding of $\overline{y}$.

Equation \ref{eq:update_rule_app} consists of the sum of a long list of weight products, each defining a possible path starting at node $j$ and terminating in the output layer. Most of these paths are not active, given that any single $T(.)$ component that is 0 will ensure that the associated path is inactive. Redefining symbols to enumerate the active paths from node $j$, created when a single training sample is passed through the network, leads to Equation 10.

\section{Additional results: varying width}
\label{app:width}

\begin{figure}[htb]
  \center{
  \includegraphics[width=0.99\textwidth]{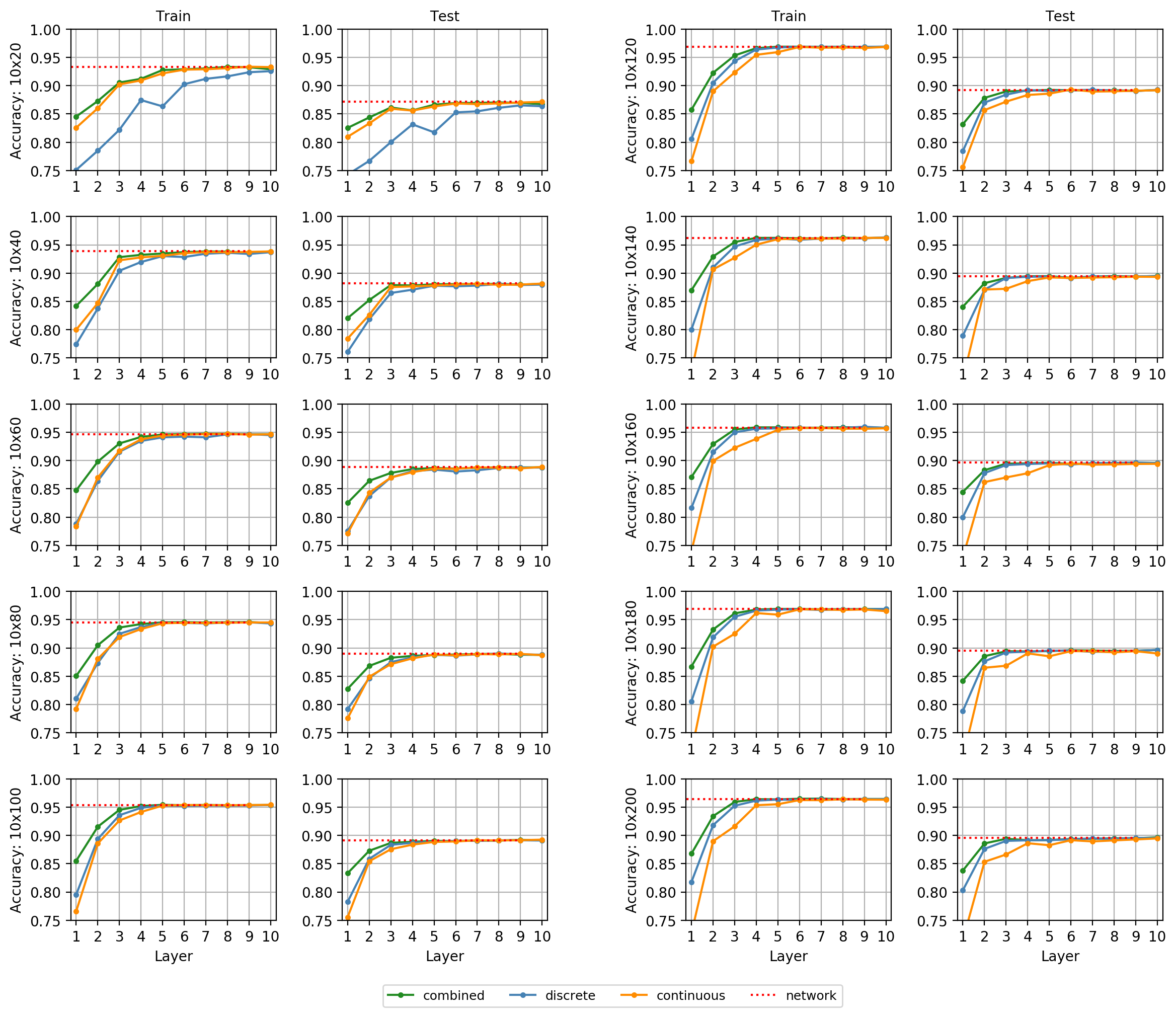}}
  \caption{Discrete, continuous and combined system train and test accuracies for networks with varied width (20-200, FMNIST).}
  \label{fig:FMINST_10_x_var}
\end{figure}

\begin{figure}[htb]
  \center{
  \includegraphics[width=0.90\textwidth]{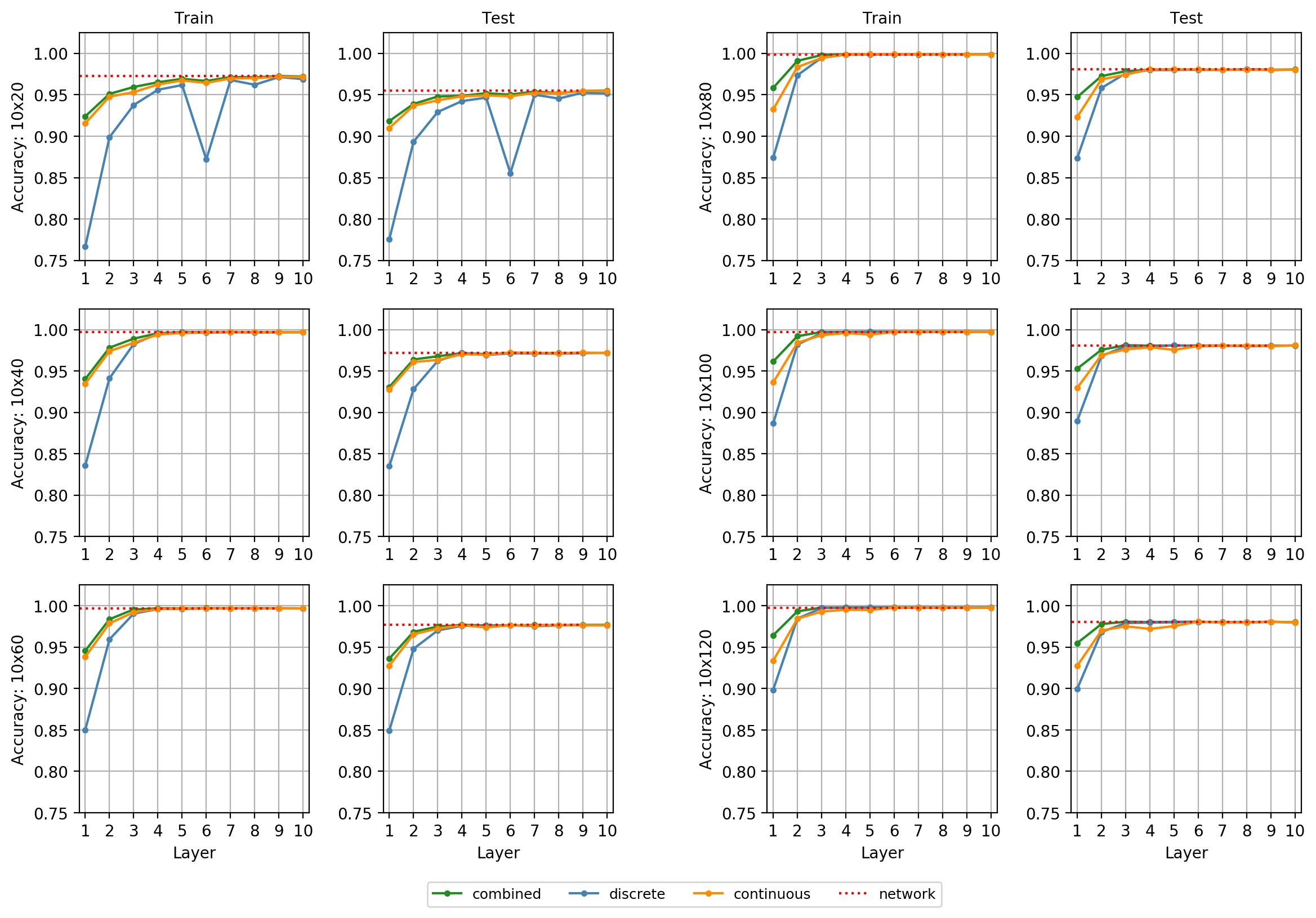}}
  \caption{Discrete, continuous and combined system train and test accuracies for networks with varied width (20-120, MNIST).}
  \label{fig:MNIST_10_x_var}
\end{figure}

In Section 5.2 we compared the performance of the discrete, continuous and combined systems for ReLU-activated networks of varying depth. 
Here we show a similiar analysis:  this time network depth is kept constant and width per layer is varied. Results are included for FMNIST (Figure \ref{fig:FMINST_10_x_var})
and MNIST (Figure \ref{fig:MNIST_10_x_var}).

\clearpage

\begin{figure}[htb]
  \center{
  \includegraphics[width=0.45\textwidth]{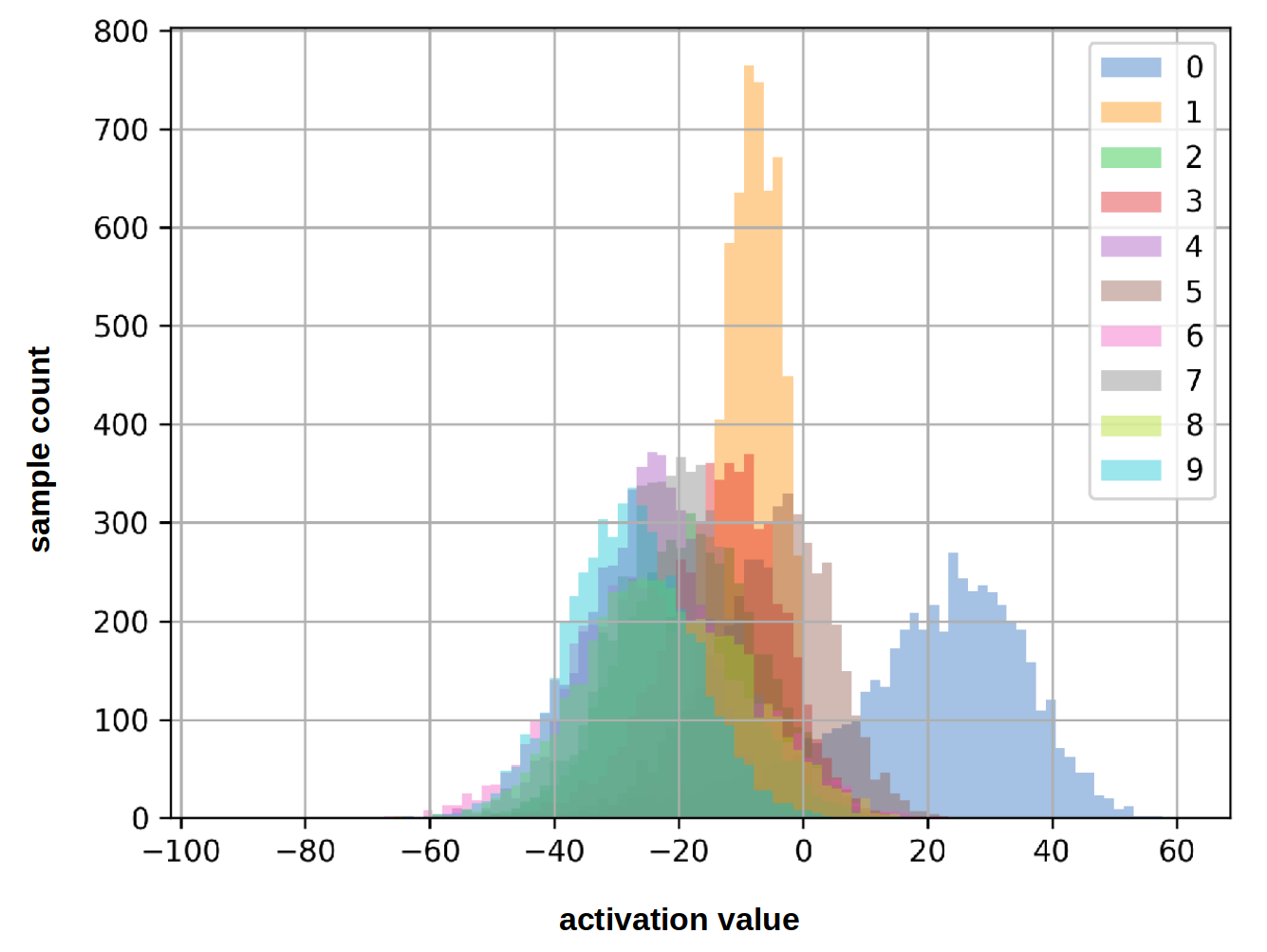}}
  \caption{Example of class activation distributions at a specific node in the first hidden layer of an MLP with ReLU activations (before the activation function is applied).
  \label{fig:node_activation_distributions}}
\end{figure}

Finally, Figure \ref{fig:node_activation_distributions} illustrates the class activation distributions at an example node:
histograms of the number of samples per class, over the range of observed activation values.
When referring to the continuous information available at each node we refer to the information captured in these class-specific distributions.

\end{document}